\documentclass[11pt]{article}
\usepackage[hyperref]{acl2023}
\usepackage{times}
\usepackage{latexsym}
\usepackage{graphicx}
\usepackage{booktabs}
\usepackage{multirow}
\usepackage{xcolor}
\usepackage{microtype}
\usepackage{url}

\title{Serialisation Strategy Matters: How FHIR Data Format\\
       Affects LLM Medication Reconciliation}

\author{
  Sanjoy Pator \\
  Independent Researcher \\
  \texttt{sanjoypatorwork@gmail.com}
}

\date{}

\begin{document}

\maketitle

\begin{abstract}
Medication reconciliation at clinical handoffs is a high-stakes, error-prone process.
Large language models are increasingly proposed to assist with this task using
FHIR-structured patient records, but a fundamental and largely unstudied variable is
how the FHIR data is serialised before being passed to the model. We present the first
systematic comparison of four FHIR serialisation strategies (Raw JSON, Markdown Table,
Clinical Narrative, and Chronological Timeline) across five open-weight models
(Phi-3.5-mini, Mistral-7B, BioMistral-7B, Llama-3.1-8B, Llama-3.3-70B) on a
controlled benchmark of 200 synthetic patients, totalling 4,000 inference runs. We
find that serialisation strategy has a large, statistically significant effect on
performance for models up to 8B parameters: Clinical Narrative outperforms Raw JSON
by up to 19 F1 points for Mistral-7B ($r = 0.617$, $p < 10^{-10}$). This advantage
reverses at 70B, where Raw JSON achieves the best mean F1 of 0.9956. In all 20
model and strategy combinations, mean precision exceeds mean recall: omission is the
dominant failure mode, with models more often missing an active medication than
fabricating one, which changes how clinical safety auditing priorities should be set. Smaller models plateau at roughly 7--10 concurrent
active medications, leaving polypharmacy patients, the patients most at risk from
reconciliation errors, systematically underserved. BioMistral-7B, a domain-pretrained
model without instruction tuning, produces zero usable output in all conditions,
showing that domain pretraining alone is not sufficient for structured extraction.
These results offer practical, evidence-based format recommendations for clinical LLM
deployment: Clinical Narrative for models up to 8B, Raw JSON for 70B and above. The
complete pipeline is reproducible on open-source tools running on an AWS
\texttt{g6e.xlarge} instance (NVIDIA L40S, 48 GB VRAM).
\end{abstract}

\section{Introduction}
\label{sec:intro}

Every hospital transfer, every discharge, every clinical handoff carries a risk: a
medication gets dropped from the list, a dose gets duplicated, or a drug a patient
has been taking for years simply disappears from the record. Medication reconciliation
is the process of comparing a patient's complete medication history against every new
prescription, and it exists specifically to catch these errors. Medication discrepancies
occur in over 60\% of patients at care transitions, and a significant fraction of these
discrepancies lead to preventable adverse events \citep{asgari2025clinicalsafety}.

Large language models are beginning to be proposed and deployed for medication
reconciliation, often as part of broader FHIR-based EHR query pipelines. The appeal
is clear: LLMs can read unstructured clinical notes, understand medication names and
dosage semantics, and produce structured output, capabilities that rule-based systems
often struggle with. But the standard format for patient health records, FHIR R4,
is deeply nested JSON built around numeric ontology codes (RxNorm, SNOMED, LOINC)
rather than readable medication names. A single patient's complete FHIR bundle can
exceed 800,000 lines of JSON \citep{schmiedmayer2025llmonfhir}, far beyond any current
context window. Before a model can reason over a patient's medication history, that
history must be preprocessed, selected, and formatted into a prompt.

Prior work has recognised the preprocessing challenge. \citet{schmiedmayer2025llmonfhir}
and \citet{schmiedmayer2024llmonfhir} use function-calling to select relevant FHIR
resource types before passing data to a model. \citet{wu2026plannerauditor} apply
deterministic auditing pipelines on top of LLM-generated FHIR plans. What none of
these works have systematically studied is a more fundamental question: given a
task-relevant subset of FHIR data, does the \textit{format} in which that data is
presented to the model matter? Should it be raw JSON? A markdown table? A clinical
narrative? A chronological timeline?

This is not a hypothetical question. Every integration pipeline that sends FHIR data
to an LLM makes an implicit serialisation choice. That choice is almost always made
based on convenience or convention, not on empirical evidence about which format
produces better outcomes. We set out to generate that evidence.

\paragraph{Contributions.} This paper makes the following contributions:

\begin{itemize}
  \item A controlled benchmark of 200 synthetic patients $\times$ 4 serialisation
        strategies $\times$ 5 open-weight models = 4,000 total inference runs, measuring
        precision, recall, and F1 for medication name extraction from FHIR bundles.

  \item Evidence that serialisation strategy has a statistically significant,
        clinically meaningful effect on performance for models up to 8B parameters
        (Mistral-7B: $r = 0.617$, $p < 10^{-10}$), and that the optimal strategy is
        model-size-dependent: Clinical Narrative is best for $\leq$8B models; Raw JSON
        is best at 70B.

  \item Evidence that omission dominates hallucination in this task. Mean precision
        exceeds mean recall in all 20 model and strategy combinations, though
        hallucinations do occur, which changes how clinical safety auditing of these
        systems should be framed.

  \item Evidence of a hard capacity ceiling for smaller models: recall degrades sharply
        for patients with more than 7--10 concurrent active medications, affecting the
        highest-risk polypharmacy patients.

  \item A reproducible open pipeline using Synthea, Ollama, and no proprietary APIs,
        running on an AWS \texttt{g6e.xlarge} instance (NVIDIA L40S, 48 GB VRAM).
\end{itemize}

The rest of this paper is structured as follows. Section~\ref{sec:related} reviews
related work; Section~\ref{sec:methods} describes the dataset, strategies, models,
and evaluation; Section~\ref{sec:results} reports findings; Section~\ref{sec:discussion}
interprets them for clinical deployment; Section~\ref{sec:limitations} discusses
limitations.

\section{Related Work}
\label{sec:related}

\subsection{LLMs for FHIR and Clinical Records}

The most direct predecessor to this work is LLMonFHIR \citep{schmiedmayer2025llmonfhir},
a Stanford physician-validated iOS application that allows patients to query their
own FHIR records using natural language. The JACC publication evaluates 210 physician-rated
responses on Synthea patients and reports median accuracy scores of 5/5, but explicitly
identifies context size as the central unresolved technical challenge: raw FHIR bundles
can reach 800,000 lines of JSON for a single patient, far exceeding current LLM context
windows. The earlier arXiv version of the same project \citep{schmiedmayer2024llmonfhir}
addresses this with GPT-4 function-calling to select relevant resource types before
sending data to the model. Our work takes a different angle: given a bounded,
task-relevant subset of FHIR resources (only MedicationRequest), how does the
\textit{format} of that subset affect model performance? This is the layer below
resource selection that prior work has not systematically examined.

\citet{wu2026plannerauditor} build a dual-agent discharge planning system using MIMIC-IV
FHIR data, where a Planner LLM generates structured discharge action plans and a
deterministic Auditor validates coverage and calibration. Their self-improvement loop
raises task coverage from 32\% to 86\%. This work is relevant because it shows that
FHIR-native LLM pipelines can reach high task coverage when structured validation is
applied, a design pattern that could be combined with our serialisation findings
for production deployment.

\subsection{Medication Extraction and FHIR-Specific Tasks}

\citet{li2024fhirgpt} introduce FHIR-GPT, which converts free-text clinical notes
and discharge summaries into FHIR MedicationStatement resources using GPT-4 and
few-shot prompting. On 3,671 medication snippets from the n2c2 2018 dataset, GPT-4
achieves over 90\% exact match, substantially outperforming prior NLP pipelines.
This is the closest prior work to our task: LLMs applied to medication-specific
FHIR resources. The key distinction is directionality. FHIR-GPT extracts from
unstructured clinical text \textit{into} FHIR format; we extract from FHIR format
\textit{into} a medication list. The inverse task uses the same data format but
requires different model capabilities: structured input comprehension rather than
structured output generation.

\subsection{Hallucination, Omission, and Clinical Safety}

\citet{asgari2025clinicalsafety} conduct the largest manual evaluation of LLM clinical
note generation to date (12,999 annotated sentences) and measure hallucination at 1.47\%
and omission at 3.45\%. They find that iterative prompt engineering can bring error
rates below human baseline and identify omission as a persistent challenge even after
optimisation. This paper establishes the methodological framing that we extend:
treating precision failures (hallucination) and recall failures (omission) as distinct
error types with different clinical consequences.

\citet{zeba2025granularfactcheck} introduce a proposition-level, LLM-free fact-checking
module that validates clinical text against EHR source data using discrete logical checks
(negation, temporal consistency, numerical comparison). Their approach achieves F1 = 0.856
for hallucination detection and demonstrates that systematic validation is tractable
without another LLM in the loop. The distinction between hallucination and omission
they formalise maps directly onto our finding that omission dominates hallucination in
this benchmark: mean precision exceeds mean recall in every model--strategy combination,
even though precision failures do occur.

\citet{kim2025medicalhallucination} evaluate medical hallucination across 11 foundation
models on 7 clinical tasks, with survey input from 70 clinicians. Their finding that
general-purpose models outperform medical-specialised ones (76.6\% vs.\ 51.3\%
hallucination-free) and that hallucination is a reasoning failure rather than a knowledge
deficit aligns closely with our BioMistral result: domain pretraining without instruction
tuning produces a model that generates no useful output on a structured extraction task.

\subsection{Prompting and Format Effects in Clinical NLP}

\citet{delaunay2025fhirprompting} evaluate three LLMs on converting Spanish clinical
reports into FHIR bundles using six prompting strategies. They find that two-step
prompting (extract then format) significantly outperforms single-step prompting, and
that temperature 0 is optimal for FHIR compliance. This is the most directly comparable
prompting study to our work. We differ in task direction (FHIR comprehension vs.\
FHIR generation), language, model family, and the specific format variable we test,
but the shared finding is that how you structure the model's task matters as much as
which model you choose.

\citet{kruse2025temporal} evaluate LLMs on longitudinal clinical summarisation and
prediction using full patient trajectories from multi-modal EHRs. A key finding is
that longer context windows improve input \textit{integration} but do not reliably
improve clinical \textit{reasoning}. This connects to our capacity ceiling result:
the failure mode for small models on complex patients is not that the records do not
fit in context, but that the models fail to enumerate all active medications in their
output even when the information is present in the input. The reasoning bottleneck, not
the context window, is the limiting factor.

\subsection{Models Evaluated}

We evaluate Phi-3.5-mini \citep{abdin2024phi3}, Mistral-7B \citep{jiang2023mistral},
BioMistral-7B \citep{labrak2024biomistral}, and models from the Llama 3 family
\citep{llama3team2024}. Synthetic patient data is generated with Synthea
\citep{walonoski2018synthea}.

\section{Methods}
\label{sec:methods}

\subsection{Dataset}

We generate a synthetic cohort of 200 patients using Synthea \cite{walonoski2018synthea},
an open-source patient simulator that produces medically realistic FHIR R4 records based
on clinical guidelines, public health statistics, and US census data. Each patient is
aged 40--75 at simulation end, has at least 10 years of medication history, and has at
least one currently active medication. Because Synthea creates one \texttt{MedicationRequest}
resource per prescription refill, a patient with a 25-year chronic disease history can
accumulate hundreds of entries for a handful of distinct drugs. No real patient data is
used at any point.

Ground truth is defined as the set of distinct medication names from all
\texttt{MedicationRequest} resources with \texttt{status = "active"} in the FHIR bundle.
Names are taken from the \texttt{medicationCodeableConcept.text} field, exactly as they
appear in the record. The ground truth set for the 200 patients ranges from 1 to 16 active
medications, with a median of 5.

\subsection{Serialisation Strategies}
\label{sec:strategies}

All four strategies receive the same FHIR bundle and produce a text string that is
embedded verbatim into the model prompt. We strip administrative fields
(\texttt{subject}, \texttt{encounter}, \texttt{requester}, \texttt{meta}, \texttt{id})
from all strategies; the model sees only clinically relevant content. Table~\ref{tab:strategy_comparison}
summarises how the four strategies differ along the dimensions that matter for model
comprehension: structural form, ordering, and how the active/historical distinction is
signalled. Appendix~\ref{app:strategy_examples} shows the complete serialised output of
each strategy for a single patient.

\begin{table*}[t]
\centering
\small
\renewcommand{\arraystretch}{1.25}
\begin{tabular}{p{2.4cm} p{3.2cm} p{3.2cm} p{3.2cm} p{3.2cm}}
\toprule
\textbf{Aspect} & \textbf{A: Raw JSON} & \textbf{B: Markdown Table} & \textbf{C: Clinical Narrative} & \textbf{D: Chrono. Timeline} \\
\midrule
Structural form &
Nested FHIR JSON, indented &
Six-column markdown table &
One sentence per medication &
Pipe-delimited timeline lines \\

Ordering &
As in the FHIR bundle, grouped by resource &
Chronological, oldest to newest &
Active medications first, then history &
Strictly chronological, oldest to newest \\

Active/historical signal &
\texttt{"status"} field inside each resource &
\texttt{active} or \texttt{completed} cell in the Status column &
Explicit heading: \textit{``Currently active medications:''} &
Inline \texttt{active} token on each line, no separation \\
\bottomrule
\end{tabular}
\caption{The four serialisation strategies compared along their structural, ordering,
  and status-signalling dimensions. Full verbatim examples of each strategy applied
  to the same patient appear in Appendix~\ref{app:strategy_examples}.}
\label{tab:strategy_comparison}
\end{table*}

\paragraph{Strategy A — Raw JSON (baseline).}
\texttt{MedicationRequest} resources are serialised as indented JSON with clinical
fields intact. Because Synthea emits one resource per refill, the full JSON bundle
for some patients exceeds the context windows of the smaller models in our evaluation
(Mistral-7B and BioMistral-7B at 32K tokens). For Strategy A only, we therefore cap
the input at 100 resources: all active medications are kept, and the remaining slots
are filled with the most recent historical records. Strategies B, C, and D apply no
such cap because their more compact representations (a flat table, per-medication
sentences, and pipe-delimited lines) fit the full history within context for every
patient in the cohort. This asymmetry is itself part of what Strategy A represents
in practice: a raw-JSON pipeline is not deployable for typical chronic-disease
histories without some form of truncation or preprocessing, so the active-first cap
is a minimally-invasive form of the filtering any real deployment would have to
perform. It is a conservative choice for Strategy A in our comparison: active items
are guaranteed to appear, so any remaining performance gap against the other
strategies reflects format effects rather than missing input.

\paragraph{Strategy B — Markdown Table.}
All \texttt{MedicationRequest} resources are flattened into a single markdown table
(columns: Medication, RxNorm code, Status, Prescribed date, Dose, Frequency), sorted
chronologically. Missing dose or frequency fields are shown as a dash. This format is
structurally explicit and human-readable while remaining machine-parseable.

\paragraph{Strategy C — Clinical Narrative.}
Each medication becomes a plain-English sentence. Active medications appear first under
a labelled heading (\textit{Currently active medications:}), followed by a separate
section for historical medications. This structure explicitly signals the task-relevant
partition, reducing the model's need to classify status from raw field values.

\paragraph{Strategy D — Chronological Timeline.}
All \texttt{MedicationRequest} resources are sorted strictly by prescription date,
oldest to newest, regardless of status. Each entry is a pipe-delimited line showing
date, status, medication name, and dosage. This format makes no attempt to distinguish
active from historical medications — the model must reason across the full temporal
sequence to determine what is currently active.

\subsection{Models}
\label{sec:models}

We evaluate five open-weight models spanning two orders of magnitude in parameter count.
Table~\ref{tab:models} summarises specifications. All models are served locally via
Ollama using 4-bit quantisation on an AWS \texttt{g6e.xlarge} instance
(NVIDIA L40S, 48 GB VRAM). No proprietary APIs are used.

\begin{table*}[t]
\centering
\small
\renewcommand{\arraystretch}{1.25}
\begin{tabular}{lrrrll}
\toprule
\textbf{Model} & \textbf{Params} & \textbf{Native Ctx} & \textbf{Runtime Ctx} & \textbf{Type} & \textbf{Quant.} \\
\midrule
Phi-3.5-mini   & 3.8B  & 128K & 64K          & Instruct & Q4\_0   \\
Mistral-7B     & 7B    & 32K  & 64K$^{*}$    & Instruct & Q4\_K\_M \\
BioMistral-7B  & 7B    & 32K  & 64K$^{*}$    & Pretrain & Q4\_K\_M \\
Llama-3.1-8B   & 8B    & 128K & 64K          & Instruct & Q4\_K\_M \\
Llama-3.3-70B  & 70B   & 128K & 32K$^{\dag}$ & Instruct & Q4\_K\_M \\
\bottomrule
\end{tabular}
\caption{Models evaluated. \textit{Native Ctx} is the model's trained maximum context;
  \textit{Runtime Ctx} is the \texttt{num\_ctx} setting passed to Ollama for our runs.
  $^{*}$Mistral-7B and BioMistral-7B were run with a runtime context exceeding their
  native 32K window; this affects 13 strategy-A patients whose serialised prompts
  exceed 32K tokens. $^{\dag}$Llama-3.3-70B was capped at 32K to keep weights plus KV
  cache within the 48 GB L40S VRAM budget. BioMistral is a further-pretrained base
  model (not instruction-tuned), included to test whether domain pretraining alone is
  sufficient. All other models are instruction-tuned.}
\label{tab:models}
\end{table*}

BioMistral~\cite{labrak2024biomistral} is further-pretrained on PubMed Central text but
is \textit{not} instruction-tuned. We include it specifically to test whether domain
pretraining alone is sufficient for a structured extraction task, given that several
recent deployments use completion-style medical models for clinical NLP tasks.

\subsection{Prompt Design}
\label{sec:prompt}

We use a single prompt template across all models and all strategies:

\begin{quote}
\small
\textit{You are a clinical assistant performing medication reconciliation.
You will be given a patient's medication history. Your task is to identify all
medications that are currently ACTIVE for this patient.
A medication is currently active if its status is ``active''. Medications with
status ``completed'', ``stopped'', ``cancelled'', or ``on-hold'' are historical
and must NOT be included in your answer.
Return your answer as a JSON array of medication names exactly as they appear
in the data. Return nothing else — no explanation, no prose, just the JSON array.
If there are no active medications, return an empty array: []}
\end{quote}

\noindent
For strategies B, C, and D, which use dashes to indicate missing dosage fields, an
additional sentence is appended clarifying that a dash represents missing data, not
inactive status. No chain-of-thought prompting or few-shot examples are used; the
prompt is designed to elicit clean structured output directly. Temperature is set to
0.0 for all runs.

\subsection{Evaluation}
\label{sec:evaluation}

We parse the model output as a JSON array of strings and compute precision,
recall, and F1 against the ground truth medication name set using \textit{exact
string match}. A predicted medication counts as correct only if it appears
verbatim in the ground truth set. This is intentionally strict: name variations
such as ``Metformin 500 MG Oral Tablet'' versus ``metformin 500mg'' count as a
miss. F1 under exact match therefore represents a lower bound on real-world
performance, where fuzzy matching would recover some of these near-misses.
If the model output cannot be parsed as a JSON array, we record
\texttt{parse\_failed=True} and assign precision = recall = F1 = 0.

We run 200 patients $\times$ 4 strategies $\times$ 5 models = 4,000 total
inference runs.

For statistical comparisons, we use the Wilcoxon signed-rank test
\citep{wilcoxon1945individual}, a non-parametric test appropriate for the
non-normal, bounded F1 distributions we observe. For within-model strategy
comparisons (Strategy A vs.\ Strategy C), p-values are Bonferroni-corrected
for four simultaneous tests per model ($k=4$). Effect sizes are reported as
$r = |Z| / \sqrt{N}$, where $N$ is the number of patient pairs, following
\citet{field2009discovering}. We use the thresholds $r \geq 0.1$ (small),
$r \geq 0.3$ (medium), and $r \geq 0.5$ (large).

\section{Results}
\label{sec:results}

\subsection{Main Results}

Table~\ref{tab:main_results} reports mean F1, precision, recall, and perfect-score
count across all 200 patients for each model--strategy combination. Two results stand out immediately. Performance varies enormously, from F1 = 0.0000 for
BioMistral on all four strategies, to F1 = 0.9956 for Llama-3.3-70B on Strategy A.
And which strategy performs best depends on model size, a finding we show is
statistically significant.

\begin{table*}[t]
\centering
\small
\resizebox{\textwidth}{!}{%
\begin{tabular}{llrrrrrrr}
\toprule
\textbf{Model} & \textbf{Strategy} & \textbf{Mean F1} & \textbf{Mean Prec.} & \textbf{Mean Rec.} & \textbf{Median F1} & \textbf{Perfect} & \textbf{Zero F1} & \textbf{Parse Fail} \\
\midrule
Phi-3.5-mini (3.8B) & A & 0.6356 & 0.7296 & 0.5932 & 0.7596 & 75  & 45  & 33 \\
Phi-3.5-mini (3.8B) & B & 0.6833 & 0.8111 & 0.6264 & 0.7500 & 64  & 25  & 16 \\
Phi-3.5-mini (3.8B) & C & \textbf{0.7008} & 0.7616 & 0.6670 & 0.8000 & 66  & 28  &  7 \\
Phi-3.5-mini (3.8B) & D & 0.6443 & 0.7703 & 0.5902 & 0.6905 & 61  & 30  & 11 \\
\midrule
Mistral-7B          & A & 0.7247 & 0.8886 & 0.6684 & 0.8889 & 87  & 20  & 11 \\
Mistral-7B          & B & 0.8753 & 0.9646 & 0.8344 & 1.0000 & 122 &  6  &  3 \\
Mistral-7B          & C & \textbf{0.9149} & 0.9319 & 0.9026 & 1.0000 & 153 & 10  &  5 \\
Mistral-7B          & D & 0.8588 & 0.9611 & 0.8100 & 1.0000 & 112 &  6  &  3 \\
\midrule
BioMistral-7B       & A & 0.0000 & 0.0000 & 0.0000 & 0.0000 &  0  & 200 & 198 \\
BioMistral-7B       & B & 0.0000 & 0.0000 & 0.0000 & 0.0000 &  0  & 200 & 197 \\
BioMistral-7B       & C & 0.0000 & 0.0000 & 0.0000 & 0.0000 &  0  & 200 & 198 \\
BioMistral-7B       & D & 0.0000 & 0.0000 & 0.0000 & 0.0000 &  0  & 200 & 198 \\
\midrule
Llama-3.1-8B        & A & 0.9180 & 0.9629 & 0.9052 & 1.0000 & 138 &  2  &  0 \\
Llama-3.1-8B        & B & 0.9250 & 0.9562 & 0.9182 & 1.0000 & 147 &  1  &  0 \\
Llama-3.1-8B        & C & \textbf{0.9471} & 0.9513 & 0.9448 & 1.0000 & 173 &  5  &  2 \\
Llama-3.1-8B        & D & 0.9228 & 0.9724 & 0.8995 & 1.0000 & 144 &  2  &  1 \\
\midrule
Llama-3.3-70B       & A & \textbf{0.9956} & 1.0000 & 0.9929 & 1.0000 & 196 &  0  &  0 \\
Llama-3.3-70B       & B & 0.9867 & 0.9900 & 0.9845 & 1.0000 & 194 &  2  &  2 \\
Llama-3.3-70B       & C & 0.9850 & 0.9850 & 0.9850 & 1.0000 & 197 &  3  &  3 \\
Llama-3.3-70B       & D & 0.8742 & 0.8750 & 0.8736 & 1.0000 & 174 & 25  &  2 \\
\bottomrule
\end{tabular}%
}
\caption{Aggregate results across all 200 patients ($n = 200$) per model and strategy.
  Strategy labels: A = Raw JSON, B = Markdown Table, C = Clinical Narrative,
  D = Chronological Timeline (full definitions in Section~\ref{sec:strategies}).
  Best strategy per model (excluding BioMistral) is in bold. Strategy C is best for
  all instruction-tuned models up to 8B; Strategy A becomes best at 70B.}
\label{tab:main_results}
\end{table*}

\subsection{The Serialisation Strategy Effect}
\label{sec:strategy_effect}

\begin{figure}[t]
  \centering
  \includegraphics[width=\columnwidth]{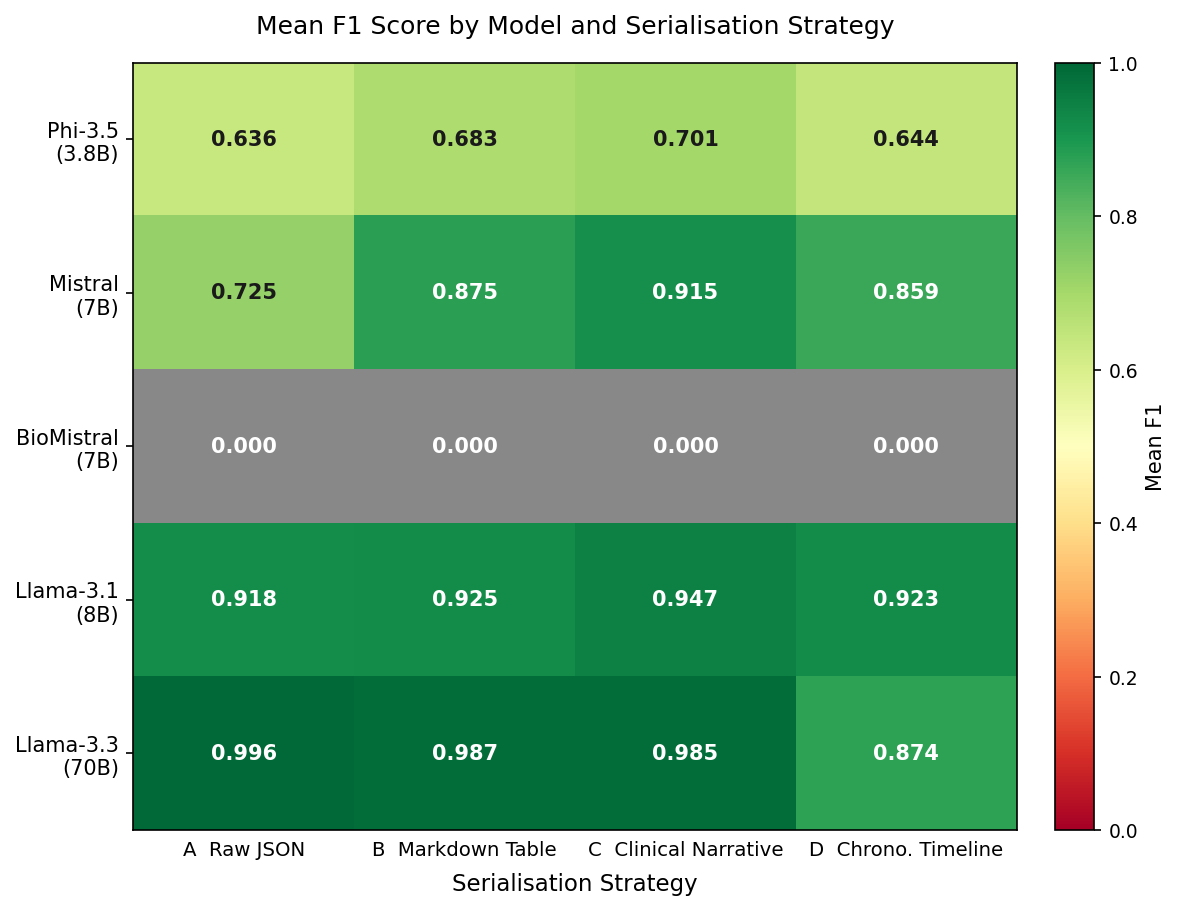}
  \caption{Mean F1 by model and serialisation strategy. Each cell shows the mean F1
           across 200 patients. Strategy C is best for $\leq$8B models; Strategy A
           is best at 70B.}
  \label{fig:heatmap}
\end{figure}

Figure~\ref{fig:heatmap} shows F1 by model and strategy as a heatmap.
For instruction-tuned models up to 8B parameters, Strategy C (Clinical Narrative)
achieves the highest mean F1 for all instruction-tuned models up to 8B: 0.7008 for Phi-3.5-mini (vs.\ 0.6356
with Strategy A), 0.9149 for Mistral-7B (vs.\ 0.7247), and 0.9471 for Llama-3.1-8B
(vs.\ 0.9180). At 70B parameters, the ranking reverses: Strategy A achieves F1 =
0.9956, which is slightly but meaningfully higher than Strategy C (0.9850). Strategy
D performs poorly at 70B (F1 = 0.8742 with 25 zero-F1 patients), suggesting that
temporal reasoning under this format is surprisingly difficult even for the largest model.

\begin{figure}[t]
  \centering
  \includegraphics[width=\columnwidth]{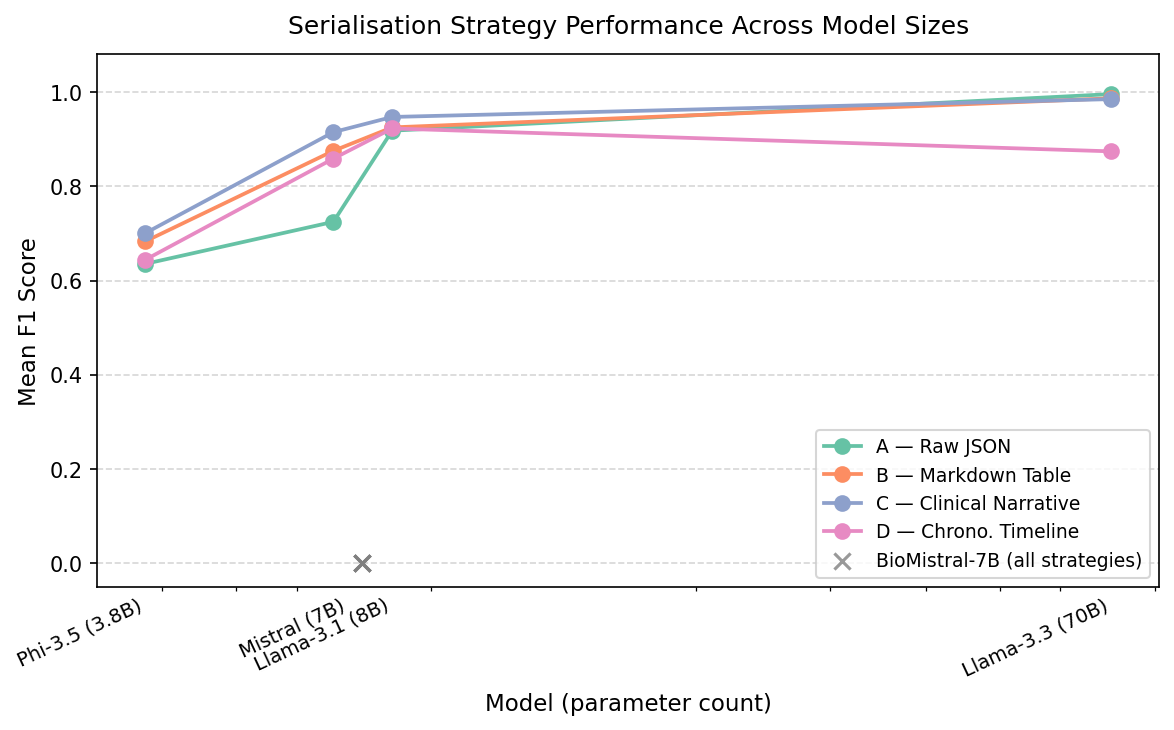}
  \caption{Mean F1 per strategy across model sizes (x-axis on log scale).
           The strategy ranking inverts between 8B and 70B.}
  \label{fig:strategy_rank}
\end{figure}

Statistical tests confirm these observations (Table~\ref{tab:stats}). For Mistral-7B,
the advantage of Strategy C over Strategy A is large and highly significant
($r = 0.617$, $p < 10^{-10}$, Bonferroni-corrected). For Llama-3.1-8B the effect is
medium ($r = 0.345$, $p = 0.011$). For Phi-3.5-mini and Llama-3.3-70B, no significant
difference is found after correction. Phi-3.5-mini is near its performance ceiling on
this task, and Llama-3.3-70B performs near-perfectly under all formats.

\begin{table}[h]
\centering
\small
\resizebox{\columnwidth}{!}{%
\begin{tabular}{llrrrl}
\toprule
\textbf{Model} & \textbf{Test} & $W$ & $p$ (corr.) & $r$ & \\
\midrule
Phi-3.5-mini   & A vs C & 3143 & 0.106  & 0.197 & ns \\
Mistral-7B     & A vs C &  948 & $< 10^{-10}$ & 0.617 & *** \\
Llama-3.1-8B   & A vs C &  859 & 0.011  & 0.345 & * \\
Llama-3.3-70B  & A vs C &   10 & 1.000  & 0.257 & ns \\
\midrule
\multicolumn{6}{l}{\textit{Cross-model comparisons on Strategy C}} \\
Mistral vs Llama-3.1-8B       & --- & 414 & 0.019       & 0.327 & *   \\
Llama-3.1-8B vs Llama-3.3-70B & --- &  22 & $< 10^{-3}$ & 0.757 & *** \\
\bottomrule
\end{tabular}%
}
\caption{Wilcoxon signed-rank tests. A vs C: Bonferroni-corrected ($k=4$).
  Significance: *** $p < 0.001$, * $p < 0.05$, ns not significant.
  $r = |Z| / \sqrt{N}$.}
\label{tab:stats}
\end{table}

\subsection{Omission Dominates Hallucination}
\label{sec:omission}

\begin{figure}[t]
  \centering
  \includegraphics[width=\columnwidth]{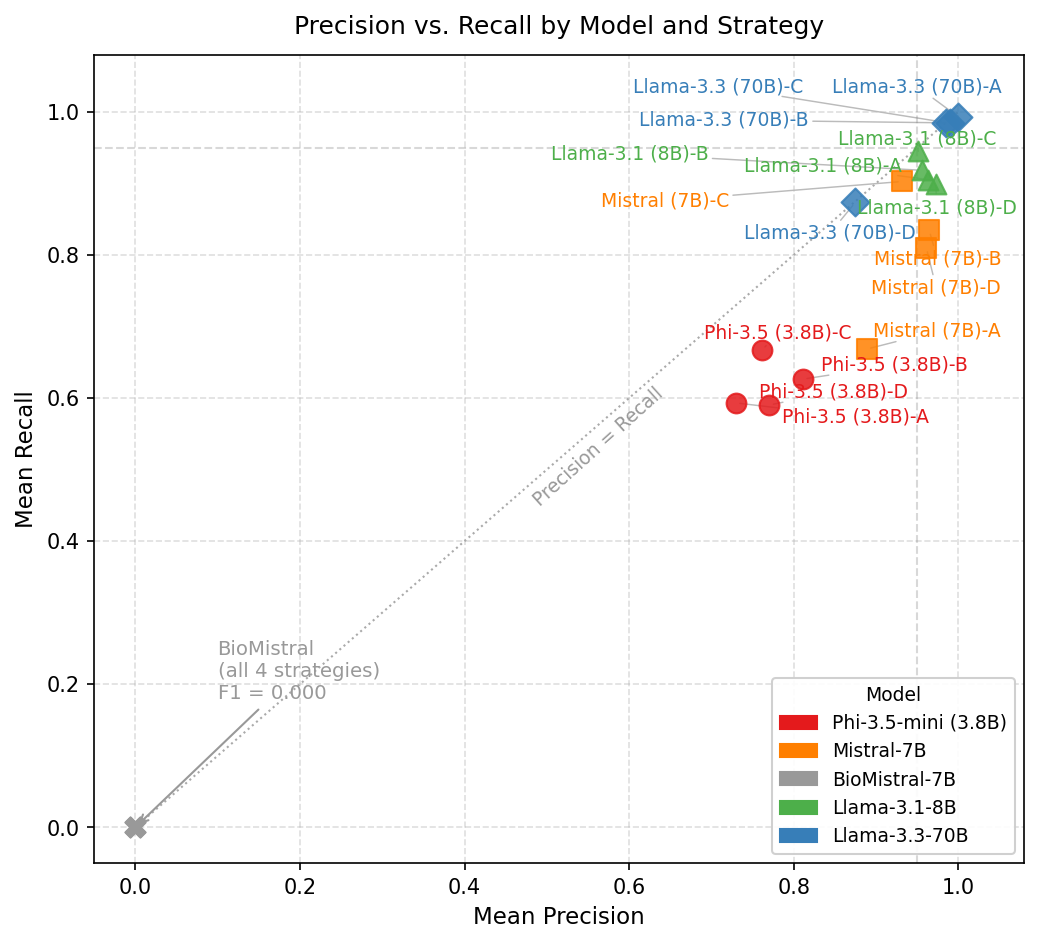}
  \caption{Precision vs.\ recall for each (model, strategy) pair. All points lie
           on or above the diagonal: omission rates exceed hallucination rates in
           every model--strategy combination.}
  \label{fig:scatter}
\end{figure}

Figure~\ref{fig:scatter} shows precision versus recall for each model--strategy
combination. In all 20 model and strategy combinations, mean precision is equal to or
higher than mean recall. Omission is the dominant failure mode: models return a subset
of the true active medication list more often than they fabricate spurious entries.
Hallucinations do occur — precision falls as low as 0.73 for Phi-3.5-mini on Raw JSON
— but at lower rates than omissions in every setting we tested.

This has direct clinical implications. The asymmetry means model output is a
\textit{partial} but \textit{higher-confidence} list of active medications. Clinicians
reviewing the output face a completeness-checking task — what did the model miss? —
rather than an equally heavy accuracy-checking task against both false positives and
false negatives.

\paragraph{What gets omitted.}
Aggregating false-negative medication names across all 200 patients reveals a clear
length bias. Omitted medications are systematically longer than retained ones: for
Mistral-7B on Clinical Narrative, 37\% of the 114 omissions have names exceeding 50
characters against a 22\% baseline among true positives; for Phi-3.5-mini on the same
strategy the gap widens to 43\% vs.\ 14\%; for Llama-3.3-70B on the Chronological
Timeline, 41\% vs.\ 22\% of 102 omissions. The most-missed single medication across
smaller models is \textit{Hydrochlorothiazide 25 MG Oral Tablet}, a first-line
antihypertensive (missed by Mistral-7B on Strategy A in 34 of 53 patients where it is
active). Top-5 omissions across models are dominated by common cardiovascular agents
(lisinopril, metoprolol succinate, simvastatin, clopidogrel), which are disproportionately
represented in the cohort and prescribed chronically with many refill entries. A
portion of Llama-3.3-70B's Strategy D precision errors are formatting variants of
correct active medications — for example \textit{lisinopril 10 MG Oral Tablet (RxNorm:
314076)} against a ground truth string without the RxNorm suffix — which under exact
matching count as both a false positive and a false negative, reinforcing that the
reported precision values are conservative lower bounds.

\subsection{Capacity Ceiling for Complex Patients}
\label{sec:ceiling}

\begin{figure}[t]
  \centering
  \includegraphics[width=\columnwidth]{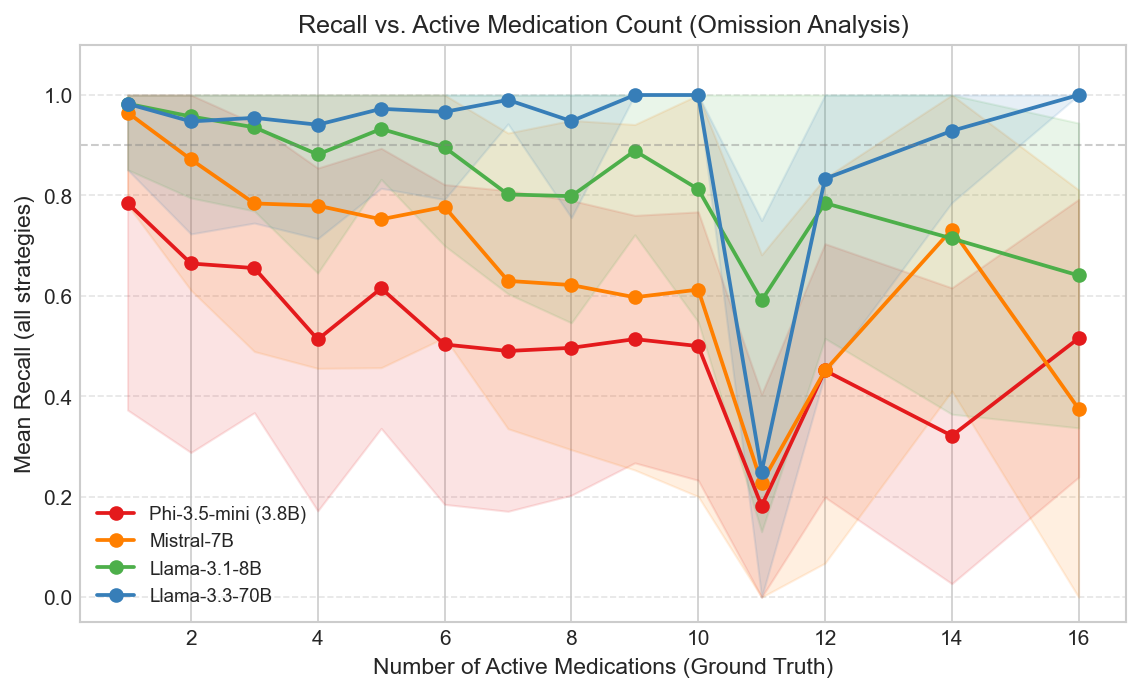}
  \caption{Mean recall by number of active medications in the ground truth. Smaller
           models plateau sharply around 7--10 active medications; Llama-3.3-70B
           remains near-perfect.}
  \label{fig:recall_gt}
\end{figure}

Figure~\ref{fig:recall_gt} shows mean recall binned by the number of active
medications in the ground truth. For models up to 8B parameters, performance degrades
sharply with increasing medication count. Mistral-7B recall falls from 0.96 for
patients with a single active medication to 0.24 for patients with 11 active
medications. Llama-3.1-8B degrades more gracefully but still drops from near-perfect
at 1--2 medications to roughly 0.80 at 10+. Llama-3.3-70B holds near-perfect recall
through 16 active medications with virtually no degradation.

\begin{figure}[t]
  \centering
  \includegraphics[width=\columnwidth]{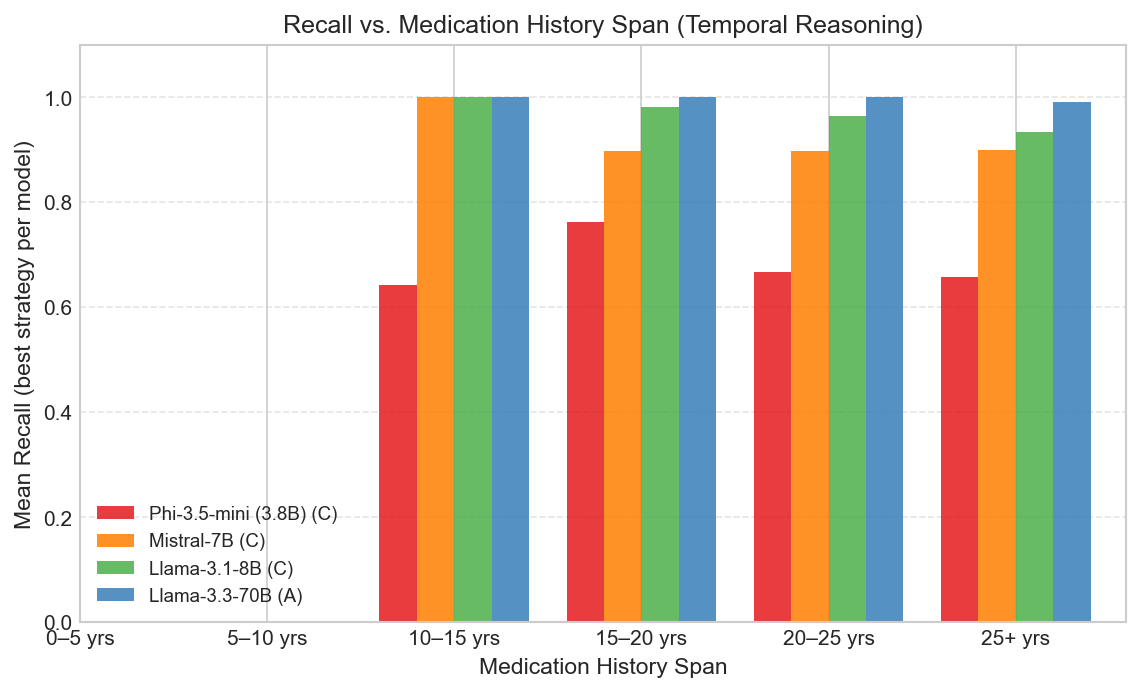}
  \caption{Mean recall by total medication history span (best strategy per model).
           Recall is flat across all history lengths, with no degradation as records
           grow longer. History span does not predict failure; only the count of
           currently active medications does (Figure~\ref{fig:recall_gt}).}
  \label{fig:recall_span}
\end{figure}

Crucially, history length (total years of medication records) does not predict
failure. Figure~\ref{fig:recall_span} confirms this: mean recall is flat across
history spans from 10 to 25+ years for every model. Only the number of
\textit{currently active} medications degrades recall. The bottleneck is output
generation capacity---listing all active medications in one response---not
input processing or reasoning over a long record.

\subsection{BioMistral: Instruction Following Failure}
\label{sec:biomistral}

BioMistral-7B achieves F1 = 0.0000 on all 200 patients across all four strategies.
Its sibling model, Mistral-7B v0.1 (same architecture, same parameter count), achieves
F1 = 0.9149 on Strategy C. The difference is not due to domain knowledge: BioMistral
was further-pretrained on PubMed Central text \textit{starting from} Mistral-7B
Instruct v0.1, which should have preserved its instruction-following capabilities in
principle.

\begin{figure}[t]
  \centering
  \includegraphics[width=\columnwidth]{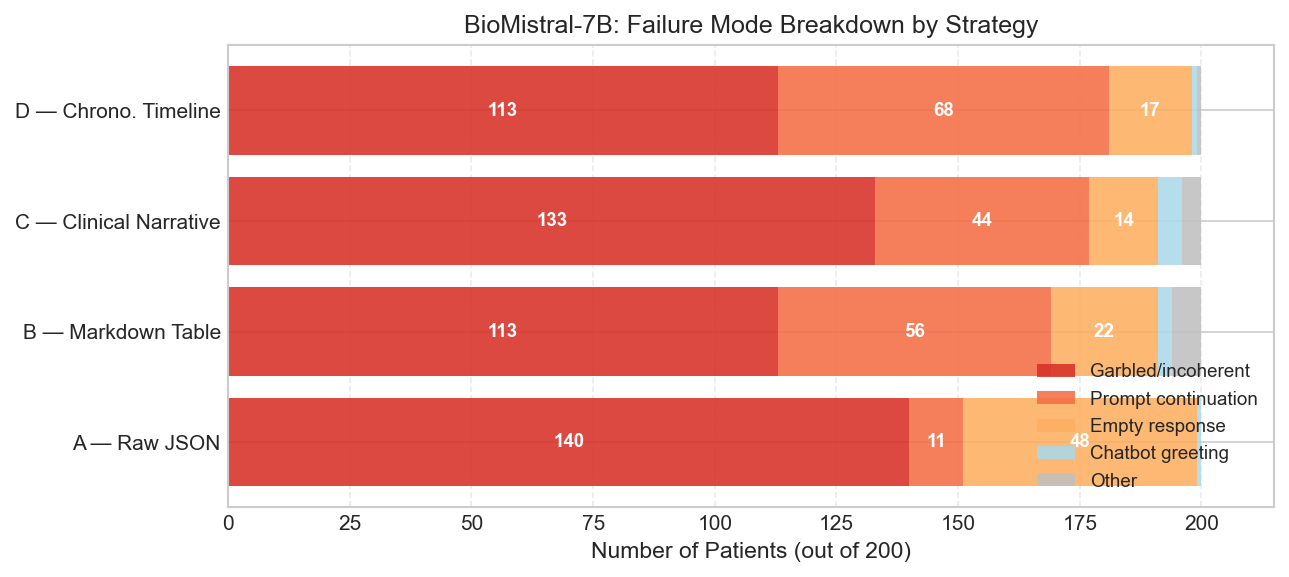}
  \caption{BioMistral-7B failure mode breakdown by strategy (200 patients each).
           Garbled or incoherent output dominates across all four strategies. Prompt
           repetition (the model echoes the system prompt instead of responding)
           accounts for 11--68 patients per strategy. No strategy produces a
           parseable JSON response.}
  \label{fig:bio_failures}
\end{figure}

Inspection of raw outputs reveals two failure modes, shown in
Figure~\ref{fig:bio_failures}: (1) garbled or incoherent token sequences
(113--140 patients depending on strategy) and (2) the model continuing the
system prompt verbatim rather than generating a response (11--68 patients).
A small number of runs produced an empty response or a generic chatbot greeting.
In no case did BioMistral return a parseable JSON array. The catastrophic
forgetting of instruction-following behaviour during domain-continued pretraining
renders the model entirely unusable for structured extraction tasks, regardless
of the input format.

\begin{figure}[t]
  \centering
  \includegraphics[width=\columnwidth]{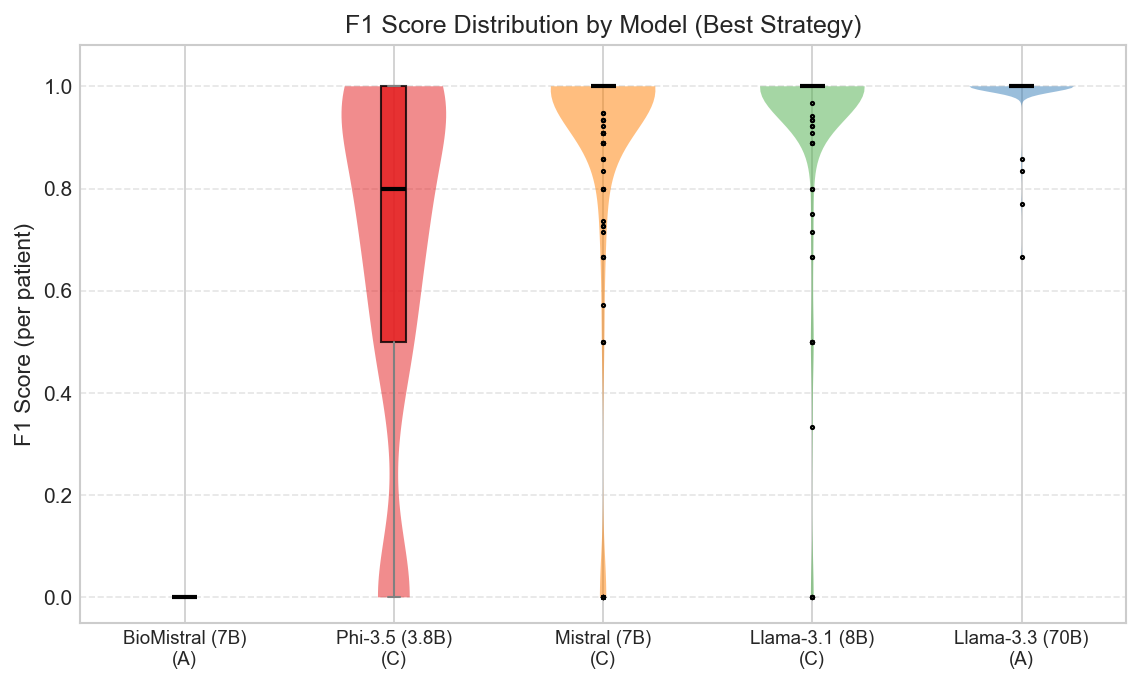}
  \caption{F1 distribution per model on best strategy. BioMistral (all strategies)
           produces only zero-F1 outputs. Llama-3.3-70B has the tightest distribution
           near 1.0.}
  \label{fig:f1_dist}
\end{figure}

Figure~\ref{fig:f1_dist} shows the F1 distribution per model on each model's best
strategy. BioMistral's distribution is a degenerate spike at zero.

\section{Discussion}
\label{sec:discussion}

\paragraph{Format is a deployment decision.}
Serialisation strategy is not a cosmetic choice. It is a performance variable with
statistically significant, clinically meaningful effect sizes. For Mistral-7B,
choosing Clinical Narrative (Strategy C) over Raw JSON (Strategy A) raises mean F1
from 0.72 to 0.91. That is a 19 percentage point gain on the same model, same hardware,
same data, with no additional training. For Llama-3.1-8B, the gain is smaller but
still significant: 0.92 to 0.95. These gains cost nothing beyond a preprocessing step.
No retraining, no fine-tuning, no prompt engineering beyond the format change itself.
Any team deploying a 7B--8B model for medication reconciliation should default to
Clinical Narrative format unless they have evidence that their specific model and data
distribution behave differently.

At 70B, the story changes. Llama-3.3-70B reaches near-perfect performance on Raw JSON
(F1 = 0.9956) and shows no significant degradation on Clinical Narrative (F1 = 0.9850),
but drops to F1 = 0.8742 on the Chronological Timeline, with 25 patients receiving zero
F1. A model of this scale appears to handle JSON structure natively without needing it
reorganised, but it still struggles with temporal inference across interleaved statuses,
even when it handles all other formats correctly. Format sensitivity does not disappear
at large scale; it shifts. Large models are more robust to structural format changes
but remain vulnerable to tasks that require specific reasoning patterns.

\paragraph{Omission defines the clinical risk profile.}
Omission is the dominant failure mode across all conditions. In all 20 model--strategy
combinations, mean precision exceeds mean recall, meaning models are substantially more
likely to miss an active medication than to fabricate one. Hallucination does still
occur, particularly for smaller models on less structured formats: Phi-3.5-mini on Raw
JSON has mean precision 0.73 (roughly 27\% of its output items are not in the ground
truth) and Mistral-7B on Raw JSON reaches 0.89. These are measurable, non-trivial error
rates. What the data shows is not the absence of hallucination but its
subordination to omission.

Two caveats sharpen this finding. First, under exact string matching, some apparent
hallucinations reflect name formatting variations rather than true fabrication: the
model outputs a real active medication with a suffix such as \textit{(RxNorm: 314076)}
or a truncated bracketed brand name, which counts as both a false positive and a false
negative. Inspection of Llama-3.3-70B's Strategy D errors shows this pattern accounts
for the majority of its 73 precision errors. Our reported precision and recall values
are therefore conservative lower bounds; fuzzy matching would raise both. Second,
the asymmetry itself is consistent across scales and strategies even at 70B, which
suggests it is a property of the task under exact-match evaluation rather than a
small-model artefact.

A model with high precision and imperfect recall produces output that can be treated
as a confirmed active subset, not a complete active list. Downstream reconciliation
workflows should be designed accordingly. Model output flags confident positives;
human review focuses on completeness rather than accuracy checking. That is a
different cognitive task, and potentially a more tractable one. Instead of verifying
every model output item is correct, the clinician's job becomes catching what the
model missed.

\paragraph{Polypharmacy patients are systematically underserved.}
Small models hit a hard capacity ceiling around 7--10 concurrent medications. For
Mistral-7B, this ceiling is stark: recall at 11 active medications falls to 0.24.
For context, roughly 30\% of adults aged 50--80 take five or more prescription
medications \citep{lantz2020polypharmacy}. Patients with 10+ active medications are
rare in the general population, but they are the patients most likely to present at
hospital handoffs and most at risk from reconciliation errors. The small-model capacity
ceiling affects precisely the population where medication reconciliation matters most.

Llama-3.3-70B does not show this ceiling. It maintains near-perfect recall through
16 active medications. The practical implication: for clinical deployments involving
complex, multi-medication patients, model scale may matter more than serialisation
choice. No format of Mistral-7B reaches Llama-3.3-70B's performance on the hardest
patients.

\paragraph{Domain pretraining without instruction tuning is insufficient.}
BioMistral produced F1 = 0.0 on all 4,000 runs assigned to it. Its sibling model,
Mistral-7B (same architecture, same parameter count), produced F1 = 0.9149 on
Strategy C. The domain knowledge was there. The instruction-following capability was
not. BioMistral was further-pretrained on 3B tokens of PubMed Central text starting
from Mistral-7B Instruct v0.1, which should have preserved instruction-following
capability in principle. It did not. The catastrophic forgetting of structured
response generation during continued pretraining rendered it entirely unusable,
regardless of input format.

This aligns with \citet{kim2025medicalhallucination}, who show that general-purpose
models outperform medical-specialised models on structured clinical tasks (76.6\% vs.\
51.3\% hallucination-free) and interpret the gap as a reasoning failure rather than a
knowledge deficit. Our result is more extreme. BioMistral does not underperform; it
produces zero useful output. For clinical NLP practitioners, the lesson is clear:
instruction-following capability is the prerequisite. A general-purpose instruction-tuned
model at the same parameter count will reliably outperform a domain-pretrained base
model on these tasks, regardless of medical vocabulary coverage.

\section{Conclusion}
\label{sec:conclusion}

Medication reconciliation is one of the highest-stakes tasks at clinical handoff points.
Missed or duplicated medications cause preventable patient harm, and every hospital
discharge or transfer is an opportunity for one to slip through. As LLMs begin to be
deployed for this task on FHIR-structured records, the question of how to present FHIR
data to a model has gone largely unexamined. This work provides the first systematic
comparison.

We evaluated 5 models across 4 serialisation strategies on 200 synthetic patients,
totalling 4,000 runs. The findings are consistent and clear. Serialisation strategy
has a statistically significant effect on performance for all instruction-tuned models
up to 8B parameters, with Clinical Narrative being the best-performing format at that
scale. The ranking flips at 70B: Raw JSON becomes the best format, and the strategy
effect loses significance. In all 20 model and strategy combinations, mean precision
exceeds mean recall: omission is the dominant failure mode, and that changes how safety
auditing of these systems should work. Small models plateau at 7--10 active medications;
polypharmacy patients remain systematically underserved below 70B scale. BioMistral's
complete failure adds a fifth finding: domain pretraining without instruction tuning is
not sufficient for structured extraction tasks.

The practical guidance from this work is straightforward. For teams using a 7B--8B
instruction-tuned model, Clinical Narrative serialisation is the right default. For
70B deployments, Raw JSON is simpler and marginally better. In both cases, the key
metric to track in production is recall, not precision. The system's completeness is
what matters clinically, not its accuracy on what it does report.

Future work should extend this benchmark to real hospital EHR data, evaluate
properly instruction-tuned biomedical models, and explore whether few-shot examples
or chain-of-thought prompting can push recall higher for complex patients. The pipeline
uses only open tools: Synthea for patient generation, Ollama for local inference, all
running on an AWS \texttt{g6e.xlarge} instance with an NVIDIA L40S (48 GB VRAM),
making the entire benchmark reproducible without proprietary APIs.

\section{Limitations}
\label{sec:limitations}

\paragraph{Synthetic data.}
All 200 patients are generated by Synthea. Synthea produces medically realistic
FHIR records, with medication names, dosages, RxNorm codes, and patient trajectories
following real clinical distributions, but it is not real patient data. Real EHR records
contain transcription errors, inconsistent formatting, missing fields, local coding
conventions, and data quality issues not present in Synthea output. We expect our F1
values to represent upper-bound estimates of performance on real hospital data for the
same task. External validation on real EHR data is the most important next step for
this work.

\paragraph{Exact string matching.}
We evaluate using exact string match against medication names as they appear in the
FHIR record. A prediction of ``metformin 500mg'' against a ground truth of ``Metformin
500 MG Oral Tablet'' counts as a miss. In real-world deployment, medication name
normalisation or fuzzy matching (e.g., RxNorm concept-based matching) would recover
many of these near-misses. Our F1 scores are therefore conservative lower bounds.
The rank ordering of models and strategies is unlikely to change under fuzzy evaluation,
but absolute values would improve.

\paragraph{English only.}
Synthea generates English-language patient records. Clinical NLP performance for
FHIR medication reconciliation in other languages, or for multilingual records, is
not addressed by this study.

\paragraph{Instruction-tuned models only (except BioMistral).}
All models tested other than BioMistral are instruction-tuned. No instruction-tuned
biomedical models (MedAlpaca, Meditron-70B, BioLlama) were included in this evaluation.
BioMistral's failure is a completion-model failure, not a biomedical-LLM failure in
general. Future work should test properly instruction-tuned medical models to assess
whether domain pretraining provides a benefit when paired with instruction fine-tuning.

\paragraph{Strategy A input cap.}
Strategy A applies an active-first, 100-entry cap on \texttt{MedicationRequest}
resources (see Section~\ref{sec:strategies}); Strategies B, C, and D do not.
This asymmetry is necessary to fit Synthea's refill-level bundles into the 32K
context windows of the 7B models, but it means Strategy A receives a mildly
preprocessed JSON rather than the raw bundle. The cap is conservative for our
comparison: it guarantees every active medication enters the model input, so any
remaining recall gap against the other strategies reflects format effects, not
missing evidence. A deployment that wants true raw JSON on a 32K-context model
would either truncate at the token level or reject long-history patients, both of
which would degrade Strategy A further.

\paragraph{Single prompt template.}
We use one prompt design across all models and strategies. We do not ablate prompt
wording, system message length, instruction phrasing, or output format constraints.
Strategy effects could interact with prompt effects in ways this study cannot disentangle.
A prompt-ablation study on a single model would help separate these contributions.

\section*{Ethics Statement}
This study uses only synthetic patient data generated by Synthea
\cite{walonoski2018synthea}. No real patient data, human subjects, or protected health
information is involved at any stage of the work. No IRB approval was required. All
models evaluated are publicly released open-weight checkpoints served locally; no
patient-derived text is sent to third-party APIs.

\section*{Data and Code Availability}
All code (serialisers, experiment runner, evaluation and analysis scripts) together
with the aggregate per-run results and the 200-patient synthetic FHIR cohort is
publicly available at
\url{https://github.com/SanjoyPator1/fhir-medrecon-serialisation}.
The repository includes the exact commands needed to regenerate the Synthea cohort,
re-run inference via Ollama on a single GPU, and reproduce every figure and statistical
test reported in this paper.

\section*{Acknowledgments}
We thank the developers of Synthea, Ollama, and the open-weight model families
(Phi, Mistral, BioMistral, Llama) whose public releases make this line of work possible
on a single-GPU reproducible budget.

\bibliography{references}

\clearpage
\onecolumn
\appendix

\section{Full Serialisation Examples}
\label{app:strategy_examples}

This appendix shows the complete serialised output of each strategy for a single patient
(Merry217 Parisian75, age 60, female; four active medications, five completed). This is
exactly the text that follows the prompt preamble and is passed to the model. The ground
truth active-medication set for this patient is:

\begin{itemize}
  \item Clopidogrel 75 MG Oral Tablet
  \item Simvastatin 20 MG Oral Tablet
  \item 24 HR metoprolol succinate 100 MG Extended Release Oral Tablet
  \item Nitroglycerin 0.4 MG/ACTUAT Mucosal Spray
\end{itemize}

\subsection{Strategy A --- Raw JSON}
\label{app:strategy_a}

\begin{quote}
\scriptsize
\begin{verbatim}
Patient: Merry217 Parisian75 | Age: 60 | Gender: female

[
  {
    "resourceType": "MedicationRequest",
    "status": "completed",
    "medicationCodeableConcept": {
      "coding": [
        {
          "system": "http://www.nlm.nih.gov/research/umls/rxnorm",
          "code": "562251",
          "display": "Amoxicillin 250 MG / Clavulanate 125 MG Oral Tablet"
        }
      ],
      "text": "Amoxicillin 250 MG / Clavulanate 125 MG Oral Tablet"
    },
    "authoredOn": "2006-02-21T14:22:22+05:30"
  },
  {
    "resourceType": "MedicationRequest",
    "status": "active",
    "medicationCodeableConcept": {
      "coding": [
        {
          "system": "http://www.nlm.nih.gov/research/umls/rxnorm",
          "code": "309362",
          "display": "Clopidogrel 75 MG Oral Tablet"
        }
      ],
      "text": "Clopidogrel 75 MG Oral Tablet"
    },
    "authoredOn": "2014-12-17T15:22:22+05:30"
  },
  ... (7 further MedicationRequest entries omitted for space:
       3 additional active medications and 4 more completed entries,
       each with the same resource shape as above)
]
\end{verbatim}
\end{quote}

\subsection{Strategy B --- Markdown Table}
\label{app:strategy_b}

\begin{quote}
\scriptsize
\begin{verbatim}
Patient: Merry217 Parisian75 | Age: 60 | Gender: female

| Medication | RxNorm | Status | Prescribed | Dose | Frequency |
| --- | --- | --- | --- | --- | --- |
| Amoxicillin 250 MG / Clavulanate 125 MG Oral Tablet | 562251 | completed | 2006-02-21 | - | - |
| Acetaminophen 325 MG Oral Tablet | 313782 | completed | 2006-04-05 | - | - |
| Clopidogrel 75 MG Oral Tablet | 309362 | active | 2014-12-17 | - | - |
| Simvastatin 20 MG Oral Tablet | 312961 | active | 2014-12-17 | - | - |
| 24 HR metoprolol succinate 100 MG Extended Release Oral Tablet | 866412 | active | 2014-12-17 | - | - |
| Nitroglycerin 0.4 MG/ACTUAT Mucosal Spray | 705129 | active | 2014-12-17 | - | - |
| Amoxicillin 250 MG / Clavulanate 125 MG Oral Tablet | 562251 | completed | 2018-01-24 | - | - |
| Amoxicillin 250 MG / Clavulanate 125 MG Oral Tablet | 562251 | completed | 2019-09-21 | - | - |
| Amoxicillin 250 MG / Clavulanate 125 MG Oral Tablet | 562251 | completed | 2025-10-17 | - | - |
\end{verbatim}
\end{quote}

\subsection{Strategy C --- Clinical Narrative}
\label{app:strategy_c}

\begin{quote}
\scriptsize
\begin{verbatim}
Patient: Merry217 Parisian75 | Age: 60 | Gender: female

Currently active medications:
  - Clopidogrel 75 MG Oral Tablet (RxNorm: 309362), prescribed on 2014-12-17,
    status: active. Dosage not recorded.
  - Simvastatin 20 MG Oral Tablet (RxNorm: 312961), prescribed on 2014-12-17,
    status: active. Dosage not recorded.
  - 24 HR metoprolol succinate 100 MG Extended Release Oral Tablet
    (RxNorm: 866412), prescribed on 2014-12-17, status: active.
    Dosage not recorded.
  - Nitroglycerin 0.4 MG/ACTUAT Mucosal Spray (RxNorm: 705129),
    prescribed on 2014-12-17, status: active. Dosage not recorded.

Medication history (no longer active):
  - Amoxicillin 250 MG / Clavulanate 125 MG Oral Tablet (RxNorm: 562251),
    prescribed on 2006-02-21, status: completed. Dosage not recorded.
  - Acetaminophen 325 MG Oral Tablet (RxNorm: 313782),
    prescribed on 2006-04-05, status: completed. Dosage not recorded.
  - Amoxicillin 250 MG / Clavulanate 125 MG Oral Tablet (RxNorm: 562251),
    prescribed on 2018-01-24, status: completed. Dosage not recorded.
  - Amoxicillin 250 MG / Clavulanate 125 MG Oral Tablet (RxNorm: 562251),
    prescribed on 2019-09-21, status: completed. Dosage not recorded.
  - Amoxicillin 250 MG / Clavulanate 125 MG Oral Tablet (RxNorm: 562251),
    prescribed on 2025-10-17, status: completed. Dosage not recorded.
\end{verbatim}
\end{quote}

\subsection{Strategy D --- Chronological Timeline}
\label{app:strategy_d}

\begin{quote}
\scriptsize
\begin{verbatim}
Patient: Merry217 Parisian75 | Age: 60 | Gender: female

Chronological medication history (oldest to newest):

2006-02-21 | completed | Amoxicillin 250 MG / Clavulanate 125 MG Oral Tablet
                        (RxNorm: 562251) | -
2006-04-05 | completed | Acetaminophen 325 MG Oral Tablet (RxNorm: 313782) | -
2014-12-17 | active    | Clopidogrel 75 MG Oral Tablet (RxNorm: 309362) | -
2014-12-17 | active    | Simvastatin 20 MG Oral Tablet (RxNorm: 312961) | -
2014-12-17 | active    | 24 HR metoprolol succinate 100 MG Extended Release
                        Oral Tablet (RxNorm: 866412) | -
2014-12-17 | active    | Nitroglycerin 0.4 MG/ACTUAT Mucosal Spray
                        (RxNorm: 705129) | -
2018-01-24 | completed | Amoxicillin 250 MG / Clavulanate 125 MG Oral Tablet
                        (RxNorm: 562251) | -
2019-09-21 | completed | Amoxicillin 250 MG / Clavulanate 125 MG Oral Tablet
                        (RxNorm: 562251) | -
2025-10-17 | completed | Amoxicillin 250 MG / Clavulanate 125 MG Oral Tablet
                        (RxNorm: 562251) | -
\end{verbatim}
\end{quote}

\noindent
In the actual experiments, lines are \textit{not} wrapped; each record is a single
line with no indentation break. Wrapping in this appendix is cosmetic, introduced
to fit the column width.

\end{document}